# The Identification of Context-Sensitive Features:
# A Formal Definition of Context for Concept Learning

**Peter Turney**

Institute for Information Technology

National Research Council Canada

Ottawa, Ontario, Canada, K1A 0R6

peter@ai.iit.nrc.ca

## Abstract

A large body of research in machine learning is concerned with supervised learning from examples. The examples are typically represented as vectors in a multi-dimensional feature space (also known as attribute-value descriptions). A teacher partitions a set of training examples into a finite number of classes. The task of the learning algorithm is to induce a concept from the training examples. In this paper, we formally distinguish three types of features: *primary, contextual,* and *irrelevant* features. We also formally define what it means for one feature to be *context-sensitive* to another feature. Context-sensitive features complicate the task of the learner and potentially impair the learner's performance. Our formal definitions make it possible for a learner to automatically identify context-sensitive features. After context-sensitive features have been identified, there are several strategies that the learner can employ for managing the features; however, a discussion of these strategies is outside of the scope of this paper. The formal definitions presented here correct a flaw in previously proposed definitions. We discuss the relationship between our work and a formal definition of relevance.

## 1  Introduction

"Context" is a widely used and ill-defined term. We are concerned here with a particular, limited type of context. In particular, we are concerned with contextual features in supervised concept learning. We assume the standard machine learning model of concept learning, where data are represented as vectors in a multi-dimensional feature space. The feature space is partitioned into a finite set of classes. In the training data, a teacher has labelled each vector with its class.

In many concept learning problems, it is possible to use common-sense knowledge to divide the features into three classes: *primary* features, *contextual* features, and *irrelevant* features. Primary features are useful for classification even when they are considered in isolation, without regard for the other features. Contextual features are useful for classification only when they are considered in combination with other features. Irrelevant features are not useful, either in isolation or in combination with other features. For example, when classifying spoken vowels, the primary features are based on the sound spectrum. The accent of the speaker is a contextual feature. The color of the speaker's hair is irrelevant.

Recent work has demonstrated that strategies for exploiting contextual information can improve the performance of machine learning algorithms (Bergadano *et al.,* 1992; Katz *et al.,* 1990; Pratt *et al.,* 1991; Turney, 1993a, 1993b; Watrous, 1991; Widmer and Kubat, 1992, 1993). A discussion of these strategies is outside of the scope of this paper. Here we discuss the problem of *identifying* contextual features, rather than the problem of *managing* contextual features. The identification problem has received surprisingly little attention, perhaps because common-sense makes the problem seem trivial. However, there may be unsuspected benefits to the automatic identification of contextual features. With automatic context identification, we may discover that context-sensitive features are typical, not exceptional. Techniques for managing contextual features may have a range of applications that have escaped our notice. We believe that a learning system that can both identify and manage context will have a substantial advantage over a system that can manage context, but requires a human operator to identify context. A precise definition of context is the first step in the construction of such a system.

In Section 2 we present the definition of relevance given by John *et al.* (1994). Our definitions employ their notation and build on their work. Section 3 reviews our previous definition of context (Turney, 1993a, 1993b) and the problem pointed out by the work of John *et al.* (1994). We introduce new definitions of primary, contextual, and irrelevant features in Section 4. These new definitions are illustrated by a simple example in Section 5. In Section 6, we discuss how the definitions may be used in practice, to identify context-sensitive features. Section 7 discusses future work and Section 8 concludes.

It is worth emphasizing that our distinction between primary and contextual is orthogonal to the distinction of John *et al.* (1994) between weak and strong relevance. However, the distinctions are closely related. As we discuss below, the primary/contextual distinction and the weakly relevant/strongly relevant distinction are *duals*. In other words, they are orthogonal but symmetric.

## 2  Definition of Relevance

We introduced a formal definition of context in our previous work on context-sensitive learning (Turney, 1993a, 1993b), but subsequent work by other researchers (John *et al.,* 1994) has exposed a flaw in our definition. John *et al.* (1994) are concerned with defining *relevant* versus *irrelevant* features, which is related to the problem of defining contextual features. We will begin by examining the definitions in John *et al.* (1994), then we will review our earlier definitions (Turney, 1993a, 1993b) and discuss the problem exposed by John *et al.* (1994). Finally, we will present new definitions that correct the problem.

The following notation comes from John *et al.* (1994). Suppose we have an $m$ dimensional feature space $F_1 \times F_2 \times \ldots \times F_m$, where $F_i$ is the domain of the $i$-th feature. Let $C$ be a finite set of classes. A training instance is a tuple $\langle \vec{X}, Y \rangle$, where $\vec{X} \in F_1 \times F_2 \times \ldots \times F_m$ and $Y \in C$. We assume that instances are sampled from $F_1 \times F_2 \times \ldots \times F_m \times C$ identically and independently with a probability distribution $p$:

$$p : F_1 \times F_2 \times \ldots \times F_m \times C \rightarrow [0,1] \qquad (1)$$

In an instance of the form $\langle \vec{X}, Y \rangle$, where $\vec{X} = \langle X_1, \ldots, X_m \rangle$, we will use $X_i$ to represent the $i$-th feature and $x_i$ to represent the value of the $i$-th feature. Similarly, $y$ is the value of $Y$.

The learning algorithm takes a sequence of instances as its input. The task of a learning algorithm is to induce a structure (such as a decision tree or a neural network). Given the feature values for a new instance, $X_1 = x_1, \ldots, X_m = x_m$, the learning algorithm can use its induced structure to predict the class of the instance, $Y = y$.

Let $S_i$ be the set of all features except $X_i$:

$$S_i = \{X_1 \ldots, X_{i-1}, X_{i+1}, \ldots, X_m\} \qquad (2)$$

Let $s_i$ be an assignment of values to all of the features in $S_i$.

**Definition 1:** The feature $X_i$ is *strongly relevant* iff there exists some $x_i$, $s_i$, and $y$ for which $p(X_i = x_i, S_i = s_i) > 0$ such that:

$$p(Y = y | X_i = x_i, S_i = s_i) \\ \neq p(Y = y | S_i = s_i) \qquad (3)$$

This definition says that $X_i$ is strongly relevant when the assertion $X_i = x_i$, in the context of the information $S_i = s_i$, provides us with additional information, which we can use to improve our prediction about the value of the class $Y$.

**Definition 2:** The feature $X_i$ is *weakly relevant* iff it is not strongly relevant and there exists a (proper) subset of features $S_i'$ of $S_i$ for which there exists some $x_i$, $s_i'$, and $y$ for which $p(X_i = x_i, S_i' = s_i') > 0$ such that:

$$p(Y = y | X_i = x_i, S_i' = s_i') \\ \neq p(Y = y | S_i' = s_i') \qquad (4)$$

Definition 2 allows for features that are relevant but redundant. For example, suppose $X_i$ is strongly relevant. Sup-pose we add a new feature $X_j$ to the set of features, where we make $X_j$ identical to $X_i$, $X_i = X_j$. Now one of $X_i$ and $X_j$ is redundant. It is easy to see that neither $X_i$ nor $X_j$ is strongly relevant, according to Definition 1. However, both are weakly relevant.

**Definition 3:** The feature $X_i$ is *irrelevant* iff it is neither strongly relevant nor weakly relevant.

John *et al.* (1994) introduced these definitions for their work in feature subset selection. A learner should use all strongly relevant features and discard all irrelevant features. Some (but not all) weakly relevant features may also be discarded. John *et al.* (1994) have demonstrated that their method of feature subset selection can improve the learner's performance.

## 3   Previous Definition of Context

In our previous definition of context (Turney, 1993a, 1993b), we did not consider the possibility of weakly relevant features. In the terminology of John *et al.* (1994), we defined a *primary* feature as a feature that is weakly relevant when $S_i' = \varnothing$. We defined a *contextual* feature as a feature that is strongly relevant, but not primary. Finally, we defined an *irrelevant* feature as a feature that is neither primary nor contextual.

In the light of the definitions given by John *et al.* (1994), it is easy to see that our definitions (Turney, 1993a, 1993b) were flawed. By our old definitions, a weakly relevant feature (i.e., a redundant relevant feature) would be (mistakenly) called irrelevant when $S_i' \neq \varnothing$. John *et al.* (1994) point out that many earlier definitions of irrelevance share this flaw. In the next section, we introduce new definitions that do not have this problem.

## 4   New Definition of Context

This section presents new definitions for primary, contextual, and irrelevant features.

**Definition 4:** Suppose that $X_i$ is either strongly relevant or weakly relevant. By definition there is a subset of features $S_i'$ of $S_i$ and an assignment of values $s_i'$ to $S_i'$ such that:

$$p(Y = y | X_i = x_i, S_i' = s_i')$$
$$\neq p(Y = y | S_i' = s_i') \tag{5}$$

$$\varnothing \subseteq S_i' \subseteq S_i \tag{6}$$

There may be several such subsets. Each such subset $S_i'$ defines a *context* in which the feature $X_i$ is (strongly or weakly) relevant. Let $\alpha_i$ be the cardinality of the smallest subset (or subsets) for which $X_i$ is relevant. Let $\beta_i$ be the cardinality of the largest subset (or subsets) for which $X_i$ is relevant. We will call $\alpha_i$ the *minimum context size* and $\beta_i$ the *maximum context size*. When $X_i$ is irrelevant, $\alpha_i$ and $\beta_i$ are undefined.

It follows from Definition 4 that $0 \leq \alpha_i \leq \beta_i \leq m - 1$. It is easy to see that $X_i$ is strongly relevant when $\beta_i = m - 1$ (Definition 1) and weakly relevant when $\beta_i < m - 1$ (Definition 2).

**Definition 5:** The feature $X_i$ is *primary* iff $\alpha_i = 0$.

A primary feature is relevant even when the context is the empty set. That is, if $X_i$ is primary, then there exists some $x_i$ and $y$ for which $p(X_i = x_i) > 0$ such that:

$$p(Y = y | X_i = x_i) \neq p(Y = y) \tag{7}$$

A primary feature is informative (about the class) when considered all by itself, without any knowledge of the values of the remaining features. Note that a primary feature may be either strongly or weakly relevant.

**Definition 6:** The feature $X_i$ is *contextual* iff $\alpha_i > 0$.

A contextual feature is only relevant when considered in some (non-empty) context; a contextual feature is irrelevant when considered in isolation. That is, if $X_i$ is a contextual feature, then, for all $x_i$ and $y$:

$$p(Y = y | X_i = x_i) = p(Y = y) \tag{8}$$

A contextual feature may be either strongly or weakly relevant.

The primary/contextual distinction and the weakly relevant/strongly relevant distinction are *duals*. In other words, they are orthogonal but symmetric.[1] This is illustrated in Table 1. The relationships among these definitions is an interesting area for future research (see Section 7).

Table 1: Duality of relevance and context-sensitivity.

| Term | Definition | Dual |
|------|-----------|------|
| strongly relevant | $\beta_i = m - 1$ | primary |
| weakly relevant | $\beta_i < m - 1$ | contextual |
| primary | $\alpha_i = 0$ | strongly relevant |
| contextual | $\alpha_i > 0$ | weakly relevant |

We have defined primary features and contextual features. Now we will define what it means for one feature to be context-sensitive to another. Let $S_{i,j}$ be the set of all features except $X_i$ and $X_j$:

$$S_{i,j} = \{X_1, \ldots, X_{i-1}, X_{i+1}, \ldots, \\ X_{j-1}, X_{j+1}, \ldots, X_m\} \quad (9)$$

Let $s_{i,j}$ be an assignment of **values to** all of the features in $S_{i,j}$. (In Equation 9, we do not mean to imply that $i < j$. The order of $i$ and $j$ does not matter.)

**Definition 7:** The feature $X_i$ is *weakly context-sensitive* to the feature $X_j$ iff there exists a subset of features $S'_{i,j}$ of $S_{i,j}$ for which there exists some $x_i$, $x_j$, $s'_{i,j}$, and $y$ for which $p(X_i = x_i, X_j = x_j, S'_{i,j} = s'_{i,j}) > 0$ such that the following two conditions hold:

$$p(Y = y | X_i = x_i, X_j = x_j, S'_{i,j} = s'_{i,j}) \\ \neq p(Y = y | X_j = x_j, S'_{i,j} = s'_{i,j}) \quad (10)$$

$$p(Y = y | X_i = x_i, X_j = x_j, S'_{i,j} = s'_{i,j}) \\ \neq p(Y = y | X_i = x_i, S'_{i,j} = s'_{i,j}) \quad (11)$$

In this definition, the first condition (Equation 10) is that feature $X_i$ must be relevant in some context that includes the feature $X_j$. The second condition (Equation 11) is that feature $X_j$ is an essential (non-redundant) component of the context. The symmetry of these two conditions implies that $X_i$ is weakly context-sensitive to $X_j$ iff $X_j$ is weakly context-sensitive to $X_i$.

Our intuition is that context-sensitivity is an asymmetric relationship. However, the two conditions in Definition 7 imply a symmetric relationship. We can get an asymmetric relationship by adding more conditions.[2]

**Definition 8:** The feature $X_i$ is *strongly context-sensitive* to the feature $X_j$ iff $X_i$ is a primary feature and $X_j$ is a contextual feature and $X_i$ is weakly context-sensitive to $X_j$.

## 5 An Illustration of the Definitions

Table 2 illustrates the above definitions. In this example, the features and the class are boolean:

$$F_1 = F_2 = F_3 = C = \{0, 1\} \quad (12)$$

Table 2 shows the probability distribution $p : F_1 \times F_2 \times F_3 \times C \to [0, 1]$.

Since $p(Y = 1) = 0.5$ and $p(Y = 1 | X_1 = 1) = 0.44$, it follows that $X_1$ is a primary feature:

$$p(Y = 1) \neq p(Y = 1 | X_1 = 1) \quad (13)$$

If the value of $X_1$ is unknown, then the class $Y$ may be either 0 or 1 with equal probability ($p(Y = 1) = 0.5$). If $X_1$ is known, then we can guess the class $Y$ with better accuracy than random guessing. If $X_1 = 1$, then $Y$ is most likely to be 0 ($p(Y = 1 | X_1 = 1) = 0.44$). If $X_1 = 0$, then $Y$ is most likely to be 1. The feature $X_1$ is primary because it gives us information about the class $Y$, even when we know nothing about the other features, $X_2$ and $X_3$.

---

[1] The symmetry is that, if we swap $\alpha_i$ and $\beta_i$, 0 and $m - 1$, and $\ldots < \ldots$ and $\ldots > \ldots$, then the equations become their duals.

[2] In informal discussions, some people have said that their intuition about context-sensitivity is better captured by Definition 7, while other people prefer Definition 8. Therefore we have provided both definitions.

Table 2: Examples of the different types of features.

| Class $Y$ | Primary $X_1$ | Contextual $X_2$ | Irrelevant $X_3$ | Probability $p$ |
|---|---|---|---|---|
| 0 | 0 | 0 | 0 | 0.03 |
| 0 | 0 | 0 | 1 | 0.03 |
| 0 | 0 | 1 | 0 | 0.08 |
| 0 | 0 | 1 | 1 | 0.08 |
| 0 | 1 | 0 | 0 | 0.07 |
| 0 | 1 | 0 | 1 | 0.07 |
| 0 | 1 | 1 | 0 | 0.07 |
| 0 | 1 | 1 | 1 | 0.07 |
| 1 | 0 | 0 | 0 | 0.07 |
| 1 | 0 | 0 | 1 | 0.07 |
| 1 | 0 | 1 | 0 | 0.07 |
| 1 | 0 | 1 | 1 | 0.07 |
| 1 | 1 | 0 | 0 | 0.03 |
| 1 | 1 | 0 | 1 | 0.03 |
| 1 | 1 | 1 | 0 | 0.08 |
| 1 | 1 | 1 | 1 | 0.08 |

Since $p(Y = y | X_2 = x_2)$ equals $p(Y = y)$ for all values $y$ and $x_2$, it follows that $X_2$ is not a primary feature. However, $X_2$ is not an irrelevant feature, since:

$$p(Y = 1 | X_1 = 1, X_2 = 1, X_3 = 1) \neq p(Y = 1 | X_1 = 1, X_3 = 1) \tag{14}$$

Therefore $X_2$ is a contextual feature. Furthermore, the primary feature $X_1$ is (strongly) context-sensitive to the contextual feature $X_2$, since:

$$p(Y = 1 | X_1 = 1, X_2 = 1) = 0.53 \tag{15}$$

$$p(Y = 1 | X_1 = 1) = 0.44 \tag{16}$$

That is, if we know only that $X_1 = 1$, then our best guess is that $Y = 0$ (by Equation 16). However, if we know that $X_1 = 1$ in the context of $X_2 = 1$, then our best bet is that $Y = 1$ (by Equation 15). The feature $X_2$ is contextual because it gives us information about the class $Y$, but only when we know the value of the primary feature $X_1$.

Finally, $X_3$ is an irrelevant feature, since, for all values $y$, $x_1$, $x_2$, and $x_3$:

$$p(Y = y | X_1 = x_1, X_2 = x_2, X_3 = x_3) = p(Y = y | X_1 = x_1, X_2 = x_2) \tag{17}$$

The feature $X_3$ gives us no information about the class, even when we know the values of the other features.

## 6  Identification of Context-Sensitive Features

In general we do not know the true probability distribution $p$. We need to estimate $p$ from the training data. Let $D$ be a sequence of training instances $\langle \vec{X}, Y \rangle$ selected from $F_1 \times F_2 \times \ldots \times F_m \times C$ identically and independently with probability distribution $p$. Let $d$ be an empirical estimate of $p$, based on the frequencies observed in the training data $D$. It is likely, due to random variation in $D$, that every feature $X_i$ will appear to be primary, if we naively apply Definition 5 to the estimate $d$. Random noise will cause the following inequality to be true, even when $X_i$ is not actually primary:

$$d(Y = y | X_i = x_i) \neq d(Y = y) \tag{18}$$

To apply the above definitions, we need to allow for the presence of noise in the training data $D$.

Let $\varepsilon$ be a small positive real number, close to zero. We may say that the feature $X_i$ appears to be primary when there is a value $x_i$ of $X_i$ and a value $y$ of $Y$, such that:

$$\left| d(Y = y | X_i = x_i) - d(Y = y) \right| > \varepsilon \tag{19}$$

This inequality allows for noise. We can adjust our sensitivity to noise by altering $\varepsilon$. When $\varepsilon$ is very close to zero, the implication is that there is little noise in the data. For a fixed sample $D$, as we increase $\varepsilon$, the number of (apparently) primary features decreases. Given a certain desired level of statistical significance (say 95%), we can use standard statistical techniques to calculate the required value of $\varepsilon$.

In addition to the problem of estimating $p$ from $D$, there is the problem of searching through all possible subsets $S_i'$ of $S_i$. In general, it is not computationally tractable to examine every possible subset of features in order to determine which features are contextual and which are primary. In practice, it will be necessary to use heuristic search procedures.

## 7    Future Work and Related Work

Perhaps the main limitation of this paper is that we introduce definitions but we do not prove any theorems. One test of the value of definitions is whether they lead to interesting theorems. Another test is whether they lead to interesting empirical results. Unfortunately, we do not yet have either theorems or empirical results. This is certainly an area for future work.

This paper does not discuss what a learner should do after it has identified context-sensitive features. This topic has been discussed elsewhere (Bergadano *et al.*, 1992; Katz *et al.*, 1990; Pratt *et al.*, 1991; Turney, 1993a, 1993b; Watrous, 1991; Widmer and Kubat, 1992, 1993). However, these authors generally assume that the distinction of contextual features from primary features takes place outside of the learning algorithm. That is, it is the responsibility of the user of the machine learning software to identify context-sensitive features. The point of this paper is that it should be possible for the learning algorithm to automatically make this distinction, without human assistance. This is interesting because many datasets may contain contextual features that humans have not yet identified as such. A learner that can automatically identify contextual features may have a significantly better performance than other learners. Verifying this hypothesis is an area for future work.

The complementary relationship between our definitions of *contextual* and *primary* (Section 4) and the definitions of *strongly relevant* and *weakly relevant* of John *et al.* (1994) (Section 2) is interesting. John *et al.* (1994) developed their definitions to address the problem of selecting a subset of features. They demonstrated that the performance of a learning algorithm can be improved by eliminating irrelevant features. Our work has a different motivation. We believe that the automatic distinction of contextual and primary features can increase the applicability of various strategies for coping with context-sensitive features. However, although the motivations are different, the duality of the definitions (Table 1) hints that there are deep links between these two problems. Exploring these links in greater detail is another topic for future research.

## 8    Conclusion

In this paper, we formally distinguished contextual features from primary features. These definitions make it possible for machine learning software to automatically distinguish primary and contextual features. After the learner has distinguished contextual features from primary features, there are several strategies that may be used for improving the robustness of the learner, but the discussion of these strategies is outside of the scope of this paper.

When $p$ is unknown, it is often possible to use background knowledge to distinguish primary, contextual, and irrelevant features. For example, consider the problem of classifying images under varying lighting conditions. In general, the lighting conditions will be contextual features, the local properties of the image will be primary features, and (say) the presence of solar flares will be irrelevant. These assertions follow from our common-sense knowledge of the world and they do not appear to require formal demonstration. This suggests that the above definitions capture an important aspect of common-sense knowledge.

## Acknowledgments

Thanks to Ronny Kohavi for drawing my attention to his own work and its impact on the definitions in my earlier work on context-sensitive learning. Thanks to Miroslav Kubat and Peter Clark for sharing their ideas about context in numerous discussions with me. Thanks to Joel Martin, Dale Schuurmans, and Ronny Kohavi for their comments on an earlier version of this paper. Thanks to two anonymous referees of the *Workshop on Learning in Context-Sensitive Domains* for their comments on an earlier version of this paper.